\documentclass[journal]{IEEEtran}

\usepackage[margin=1in]{geometry}
\usepackage{amsmath,amsthm,amssymb,hyperref}
\usepackage{cite}
\usepackage{fancyhdr}
\usepackage{graphicx}
\usepackage{gensymb}
\usepackage{comment}
\usepackage{bm}      
\usepackage{tabularx} 
\usepackage{makecell} % Add this package to enable \makecell

\usepackage{algorithmic}
\usepackage{textcomp}
\usepackage{hyperref}
\usepackage{wrapfig}
\usepackage{xcolor}
\usepackage{soul}

\usepackage{adjustbox}
\usepackage{lipsum}
\usepackage{caption}
\usepackage{subcaption}
\usepackage{multirow}
\usepackage{array}

\title{On Neural Inertial Classification Networks for Pedestrian Activity Recognition }

\author{
    Zeev Yampolsky, Ofir Kruzel, Victoria Khalfin Fekson, Itzik Klein
    
    The Hatter Department of Marine Technologies, University of Haifa, Israel
}

\date{January 2025}

\begin{document}

\maketitle

\begin{abstract}
Inertial sensors are crucial for recognizing pedestrian activity. Recent advances in deep learning have greatly improved inertial sensing performance and robustness. Different domains and platforms use deep-learning techniques to enhance network performance, but there is no common benchmark. The latter is crucial for fair comparison and evaluation within a standardized framework. The aim of this paper is to fill this gap by defining and analyzing ten data-driven techniques for improving neural inertial classification networks. In order to accomplish this, we focused on three aspects of neural networks: network architecture, data augmentation, and data preprocessing. The experiments were conducted across four datasets collected from 78 participants. In total, over 936 minutes of inertial data sampled between 50-200Hz were analyzed. Data augmentation through rotation and multi-head architecture consistently yields the most significant improvements. Additionally, this study outlines benchmarking strategies for enhancing neural inertial classification networks.
\end{abstract}

\IEEEpeerreviewmaketitle

\section{Introduction}
A vital part of modern technology is the use of inertial sensors, which power many of the key features we use in our daily lives. This technology safeguards our health, enables the next generation of robotics and autonomous vehicles, and provides accurate indoor navigation. Accelerometers measure the specific force vector and gyroscopes measure the angular velocity vector, enabling motion and orientation monitoring. An inertial measurement unit (IMU) integrates three orthogonal accelerometers and gyroscopes \cite{groves}. The application of inertial sensors span a wide range of domains and platforms. These include navigation, motion tracking, virtual reality, robotics, antenna pointing, pedestrian dead reckoning, biomedical and healthcare applications, sports, agriculture, the internet of things, wearable devices, and autonomous vehicles. 
\noindent 
By processing large datasets automatically and extracting features, deep learning (DL) has achieved breakthroughs across domains such as computer vision, natural language processing, and time-series analysis, surpassing model-based and classical machine learning methods \cite{sarker2021deep, malhotra2023recent}. Recently, DL has emerged as a powerful method for addressing challenges in inertial sensing and fusion. It has demonstrated significant improvements in sensor calibration, denoising, and navigation across diverse platforms. The use of deep learning models, including convolutional neural networks (CNNs), recurrent neural networks (RNNs), as well as attention-based architectures, has demonstrated exceptional success in capturing complex nonlinear patterns from inertial data outperforming other methods \cite {cohen2024inertial, klein2022data, chen2024deep}. \\
\noindent 
One of the first DL implementations with inertial data was in activity recognition.  In human activity recognition (HAR) and smartphone location recognition (SLR), sensory data is used to identify the user activity. HAR has numerous applications including gesture recognition~\cite{ling2021comparative}, gait analysis~\cite{mason2023wearables}, healthcare~\cite{saeed2022intelligent}, and indoor navigation~\cite{shavit2021boosting}. Due to its wide applicability it has been addressed and surveyed extensively in the literature~\cite{qiu2022multi, gu2021survey, chen2021deep}. In SLR, the user's actions are reflected in changes in the smartphone's location~\cite{klein2019smartphone}. Consider, for instance, a pedestrian who places their smartphone in their right front pocket when walking (pocket mode). When walking, the pedestrian can hold the phone while walking (swing mode) and then remove it to send a text (texting mode). HAR and SLR play a particularly significant role in navigation solutions that rely solely on smartphone inertial sensors. HAR and SLR have been shown to improve pedestrian dead reckoning (PDR) accuracy when used as a prior~\cite{yang2016pacp, elhoushi2015online, klein2018pedestrian, KLEIN2025104077}. Other navigation-related problems, such as step length estimation~\cite{chen2018ionet, wang2019pedestrian, klein2020stepnet} and adaptive attitude and heading reference system (AHRS)~\cite{vertzberger2021attitude,vertzberger2022adaptive}, were also improved with the addition of SLR module.\\
\noindent However, despite the advancements in HAR and SLR several challenges remain. One of the key obstacles lies in the variability of sensor noise, body placement of devices, and environmental conditions. This can introduce inaccuracies in inertial measurements. To generalize effectively across different conditions and users, deep neural networks often require innovative training methodologies, data augmentation strategies, and optimization techniques due to the computational cost and data-hungry nature~\cite{bengio2013deep, kawaguchi2017generalization}
. Additionally, deep-learning techniques have been used to enhance network performance across different domains and platforms, but there are no common benchmarks. The latter is critical for fair comparison and evaluation in a standardized framework as well as development in the field.\\
%This paper
\noindent 
Recently, 13 data-driven techniques for improving neural inertial regression networks where evaluated  across six real-world recorded datasets (with eight sub-datasets) collected from diverse platforms including quadrotors, doors, pedestrians, and mobile robots~\cite{fekson2025enhancement}. Motivated from this work, we examine the best proven techniques on neural inertial classification networks for pedestrian activity recognition. In order to accomplish this, we distinguish between three major approaches commonly used in data-driven theory: network architectural design, data augmentation, and data preprocessing. As part of network architectural design, we examine multi-head architectures. Data augmentation techniques include rotation and additive noise. Our data preprocessing methods include denoising of the inertial signals.\\
\noindent 
We evaluate these techniques across four real-world recorded datasets. One SLR dataset and three HAR datasets. These four datasets contain data  collected from 78 participants. In total, we used 936 minutes of recorded inertial data sampled between 50-200Hz.
Research on this topic provides a comprehensive evaluation of multiple strategies across various datasets, unlike many studies that examine isolated approaches to a specific application. In this way, a broader perspective can be gained on which methods consistently improve model accuracy and which may be more effective in certain scenarios. Through the use of these techniques in various real-world contexts, this paper provides practical insight into the implementation of neural networks in inertial classification networks for pedestrian activity recognition.\\
Among all the techniques evaluated, data augmentation through rotation and multi-head  architectures emerged as the most consistently effective methods to improve the performance of neural inertial classification networks. \\
\noindent The rest of the paper is organized as follows: Section \ref{sec:methodology} gives our methodology including the baseline network with the ten data-driven techniques used in this paper. Section \ref{sec:res} presents our results and Section \ref{sec:conclusions} gives the conclusions of this study. 
\section{Methodology} \label{sec:methodology}
\noindent 
In this section, we describe the methodological framework used in this study. As a starting point, it presents the baseline network architecture for evaluating data-driven paradigms. The next section discusses principles of network design, data augmentation, and data preprocessing. 
\subsection {Baseline Architecture} \label{sec:baseline_net}
\begin{figure*}[!htb]
    \centering
    \includegraphics[width=\textwidth, height=2cm]{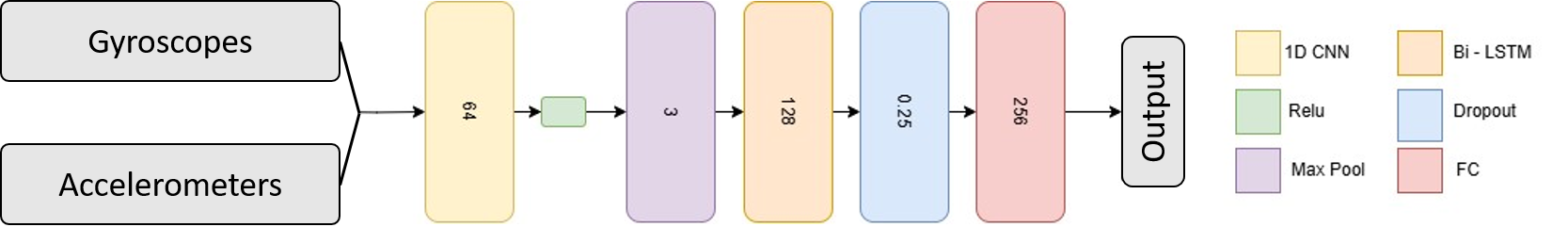}
    \caption{Baseline network architecture. The output differs between each task and dataset.}
    \label{fig:network}
\end{figure*}

\noindent Our baseline network architecture was inspired by the network presented in \cite{silva2019end} and later implemented in~\cite{fekson2025enhancement}. This architecture integrates convolutional neural network (CNN) layers, bidirectional long short-term memory (Bi-LSTM) layer, and a fully connected (FC) layer, as presented in Figure \ref{fig:network}.
The network input is a time-series signal of the inertial readings:
\begin{equation}
\textbf{x}_t = [f_x, f_y, f_z,\omega_x,\omega_y,\omega_z]^{T}\in\mathbb{R}^6
\end {equation}
where $\textbf{f}=[f_x, f_y, f_z]^{T}\in\mathbb{R}^3$ is the specific force vector as measured by the accelerometers, $\boldsymbol{\omega} = [\omega_x,\omega_y,\omega_z]^{T}\in\mathbb{R}^3$ is the angular velocity vector measured by the gyroscopes, and \(t\) is the time index. Note that the inertial measurements are expressed in the sensor coordinate frame. It is omitted from the mathematical notation for brevity.\\
\noindent The input signal is passed through one layer of 1D CNN layer with \(F=64\) filters, a kernel size \(k = 5\), and stride \(s=1\). The convolution operation for each feature map \(y_f(t)\) is given by:
\begin{equation}
\hspace{-0.2cm} % Moves the equation 1 cm to the left
y_f(t) = \mathrm{ReLU}\left(\sum_{c=1}^6 \sum_{i=1}^k w_f^{(c,i)} x_c(t+i-1) + b_f\right)
\end {equation}
where \(w_f^{(c,i)}\) are the learnable weights for filter \(f\), channel \(c\), and offset \(i\), \(b_f\) is the bias term, and \(x_c(t+i-1)\) is the input for channel \(c\) at time step \(t\).
The ReLU is a nonlinear activation function defined by:
\begin{equation}
\mathrm{ReLU}(x) = \max(0, x)
\end {equation}
where $x$ is the input to the activation function. 
The output from the convolution layer is then passed into a max pooling layer to reduce its temporal dimension:
\begin{equation}
y_f^{\text{p}}(t) = \max_{i \in (0, d)} y_f(t+i),
\end {equation}
where \(y_f^{\text{p}}(t)\) is the output of the max pooling layer at time step \(t\), \(y_f(t+i)\) is the value of the feature map \(f\) at time \(t+i\) before pooling, and \(d = 3\) is the pooling depth, which defines the size of the window used for pooling.  \\
\noindent The output of the pooling layer is fed into a Bi-LSTM layer. It processes the sequence in both forward and backward directions, producing hidden states for each time step:
\begin{equation}
\overrightarrow{\bm{h_t}} = \text{LSTM}({\bm{x_t}}, \overrightarrow{\bm{h_{t-1}}}), \quad
\overleftarrow{\bm{h_t}} = \text{LSTM}({\bm{x_t}, \overleftarrow{{\bm{h_{t+1}}}}})
\end {equation}
where \( \overrightarrow{\bm{h_t}} \) is the hidden state at time step \( t \) computed by the forward LSTM layer based on the current input \( x_t \) and the previous forward hidden state \( \overrightarrow{\bm{h_{t-1}}} \), and \( \overleftarrow{\bm{h_t}} \) is the hidden state at time step \( t \) computed by the backward LSTM layer using the current input \( x_t \) and the next backward hidden state \( \overleftarrow{\bm{h_{t+1}}} \).
The concatenated hidden states are then passed through a dropout layer to prevent overfitting during training~\cite{srivastava2014dropout}. 
\begin{equation}
\bm{h_t^{\text{dropout}}} = \bm{h_t} \odot \bm{r}, \quad \bm{r} \sim \text{Bernoulli}(1-p)
\end {equation}
where \( \bm{h_t^{\text{dropout}}} \) is the output after applying dropout, \( \bm{r} \) is a binary mask sampled from a Bernoulli distribution with probability \( 1-p \), where \( p = 0.25\) is the dropout rate, and \(\odot\) denotes element-wise multiplication.
The output from the dropout layer is then passed through a FC layer with 256 neurons:
\begin{equation}
\bm{y_{\text{FC}}} = \bm{h_t^{\text{dropout}}}\mathbf{W}_{\text{FC}} + \bm{b}_{\text{FC}} 
\end{equation}
where \( \mathbf{W}_{\text{FC}} \) is the weight matrix and \( \bm{b}_{\text{FC}} \) is the bias vector.\\
\noindent The Adam optimizer \cite{kingma2014adam} was employed with a learning rate of 0.001. The training was done on a single NVIDIA GeForce RTX 4090 GPU with a batch size of 64 samples. The number of epochs varied for each dataset, depending on factors such as convergence and running time. This will be discussed in Section \ref{sec:datasets} for each dataset. 
\subsection{Data-driven perspectives} \label{sec:paradigms}
\noindent This section addresses the approaches used in the current research to process and optimize inertial classification networks for pedestrian activity recognition. 
\subsubsection{Network Architectural Design}
\begin{figure*}[!htb]
    \centering
    \includegraphics[width=\textwidth]{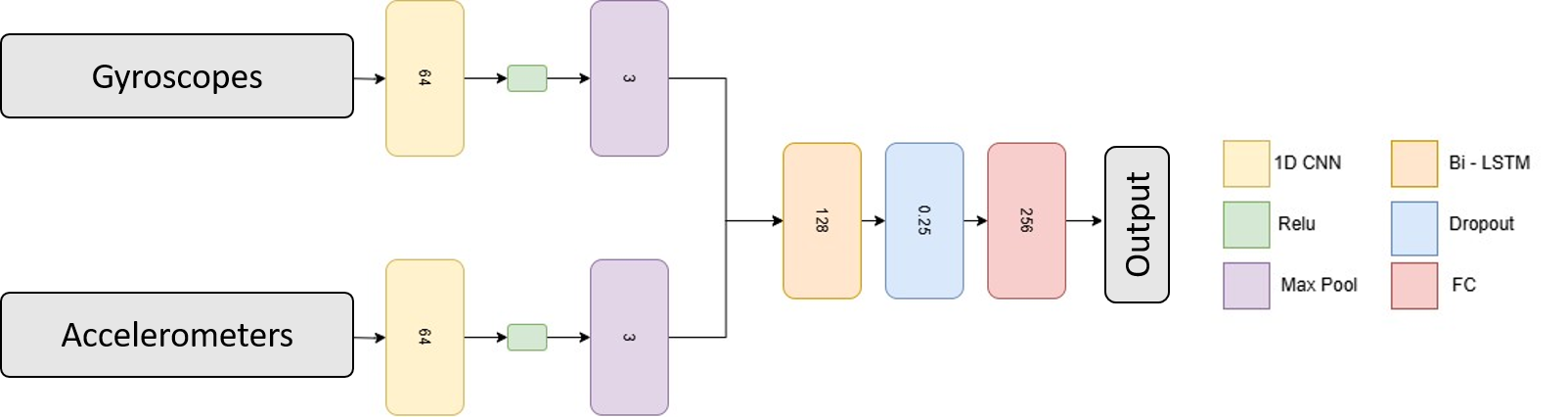}
    \caption{Head2 architecture. One head receives the accelerometer readings and the other receives the gyrosopce readings.}
    \label{fig:two_head}
\end{figure*}

\begin{figure*}[!htb]
    \centering
    \includegraphics[width=\textwidth, height=4.5cm]{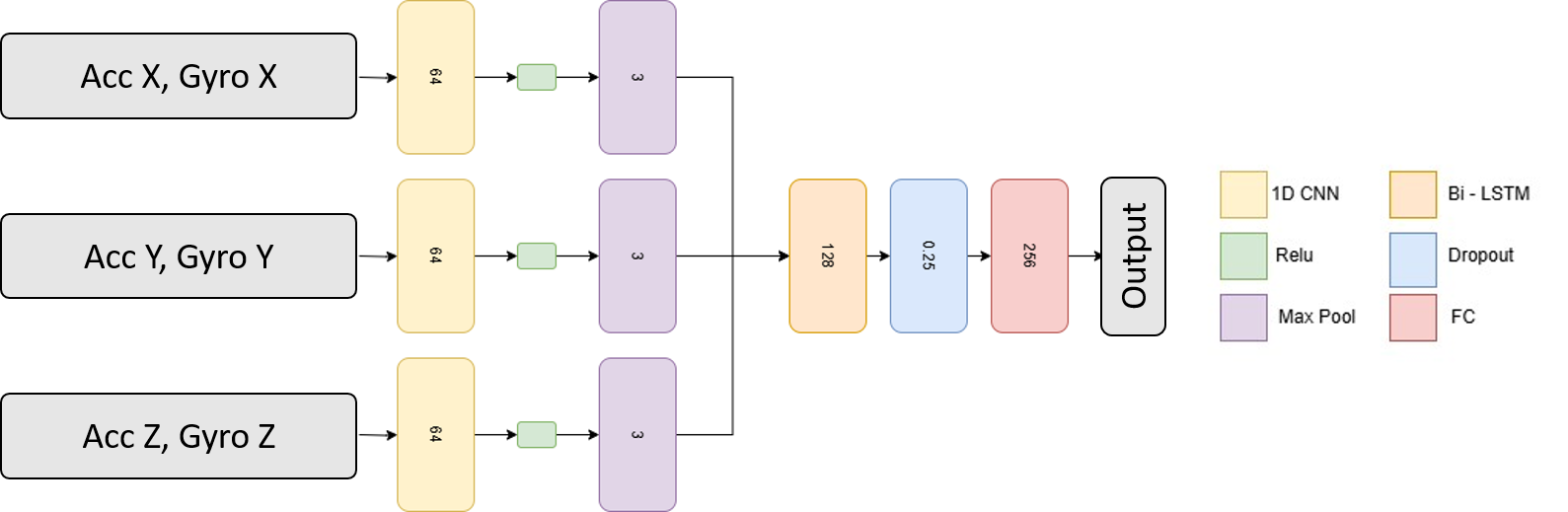}
    \caption{Head3 architecture. Each head receives the accelerometer and gyroscope readings along the x,y, and z axes.}
    \label{fig:three_head}
\end{figure*}

\paragraph{\textbf{Multi-Head Network}} \label{sec:multi_head}
Various deep learning architectures have been applied to inertial data processing tasks. Among these architectures, some utilize a single-head approach where both accelerometer and gyroscope data are passed through a single processing unit \cite{herath2020ronin,esfahani2019aboldeepio, shavit2021boosting}. Alternatively, some works used a two-head architecture, where accelerometer data is processed separately from gyroscope data \cite{silva2019end,liu2023smartphone}.  We offer two different multi-head architectures for the evaluation:
\begin{itemize}
\item \textbf{Head2} The architecture consists of two heads, one for the accelerometer readings and one for the gyroscopes, as shown in Figure \ref{fig:two_head}. 
\item \textbf{Head3}  The architecture consists of three heads, one for each inertial axis. Essentially, we integrate the accelerometer and gyroscope on the x-axis in a single head, and do the same for the y- and z-axes, as presented in Figure \ref{fig:three_head}.
\end{itemize}
\subsubsection{Data Augmentation} \label{subsec:aug}
\noindent To achieve good performance, deep learning networks require a substantial amount of data. In spite of this, collecting extensive datasets is often challenging.  To address this problem, multiple research fields have developed techniques for adding additional training data \cite {he2016deep},
For inertial data, some of the augmentation techniques include rotation, permutation, scaling, cropping, additive bias, and additive noise \cite{um2017data}. 
In this work, we examine two augmentation techniques: data rotation and additive noise. We examine each of the two separately by adding the new generated data to the existing training set.
\paragraph{\textbf{Rotation}}
Rotation augmentation enhances time series datasets by applying rotational transformations to the input data. This approach maintains the relationships between features while introducing variability and improving model generalization. 
The rotation transformation is defined by:
\begin{align}
    \tilde{\bm{x}} &= \mathbf{T} \cdot \bm{x} \label{eq:rot} 
\end{align}
where $\bm{x}$ is a sample of the inertial measurements vector, $\mathbf{T}$ is the orthonormal transformation matrix, and $\tilde{\bm{x}}$ is the rotated vector.\\
\noindent
We examine four types of rotation:
\begin{enumerate}
    \item $x$-axis rotation. The accelerometer and gyroscope readings are rotated by and angle of $\pi/6$ around the $x$-axis:
    \begin{equation}\label{eq:T1}
    \mathbf{T}_1 = \begin{bmatrix}
    1 & 0 & 0 \\
    0 & \cos\left(\frac{\pi}{6}\right) & \sin\left(\frac{\pi}{6}\right) \\
    0 & -\sin\left(\frac{\pi}{6}\right) & \cos\left(\frac{\pi}{6}\right)
    \end{bmatrix}
    \end{equation}
    \item $y$-axis rotation. The accelerometer and gyroscope readings are rotated by and angle of $\pi/6$ around the $y$-axis:
    \begin{equation}\label{eq:T2}
    \mathbf{T}_2 = \begin{bmatrix}
        \cos\left(\frac{\pi}{6}\right) & 0 & -\sin\left(\frac{\pi}{6}\right) \\
        0 & 1 & 0 \\
        \sin\left(\frac{\pi}{6}\right) & 0 & \cos\left(\frac{\pi}{6}\right)
        \end{bmatrix}
    \end{equation}
    \item $z$-axis rotation. The accelerometer and gyroscope readings are rotated by and angle of $\pi/6$ around the $z$-axis:
    \begin{equation} \label{eq:T3}
    \mathbf{T}_3  = \begin{bmatrix}
    \cos\left(\frac{\pi}{6}\right) & \sin\left(\frac{\pi}{6}\right) & 0 \\
    -\sin\left(\frac{\pi}{6}\right) & \cos\left(\frac{\pi}{6}\right) & 0 \\
    0 & 0 & 1
    \end{bmatrix}
    \end{equation}
    \item All axes rotation. Three additional datasets were generated by passing the accelerometer and gyroscope readings through \eqref{eq:T1}-\eqref{eq:T3}.
\end{enumerate}
\paragraph{\textbf{Additive  noise}}
An often-used data augmentation technique, noise addition adds random perturbations to data in order to enhance the robustness of machine learning models. For inertial time series, noise simulates real-world factors such as sensor inaccuracy, environmental disturbances, or random fluctuations. The noise is drawn from a Gaussian distribution with a mean of zero and a standard deviation specific to the dataset. The noise addition process can be expressed mathematically as follows:
\begin{align}
    \tilde{\bm{x}} &= \bm{x} + \mathcal{N}(0, \sigma^2) \label{eq:aug_add_noise}
\end{align}
where \( \bm{x} \) represents the original input sample, \( \mathcal{N}(0, \sigma^2) \) denotes the Gaussian noise with zero mean and variance \( \sigma^2 \), and \( \tilde{\bm{x}} \) is the augmented sample with the added noise.
\subsubsection{Data Preprocessing Techniques - Denoising} 
Denoising is commonly explored to assess its potential impact on improving the quality of measurements and reducing the influence of sensor noise and increasing the signal to noise ratio~\cite {khaddour2021survey}.  One of the common signal processing approach for denoising is the moving average (MA) filter \cite{engelsman2023data}. MA techniques serve as efficient smoothing filters, leveraging errors (residuals) from previous forecasts \cite{gonzalez2018statistical}. \\
\noindent The moving average filter is defined by:
\begin{equation}
\tilde{x}_t = \frac{1}{n}\sum^{n-1}_{i=0} x_{t+i} \label{eq:ma_eq}
\end{equation}
where \( n \) is the window size, \( x_{t+i} \) is the accelerometer or gyroscope measurements at each time step within the window, and \(\tilde{x}_t\) is the denoised value at time \(t\).\\
\noindent  We examined three different window sizes for the moving average filter: 10, 25, and 50 samples.
\subsection {Datasets} \label{sec:datasets}
\noindent We used four different datasets collected from 78 participants. In total, we used 936 minutes of recorded inertial data sampled between 50-200Hz.
\subsubsection{RIDI Dataset} \label{ridi}
\begin{figure}[h]
    \centering
    \includegraphics[width=\columnwidth]{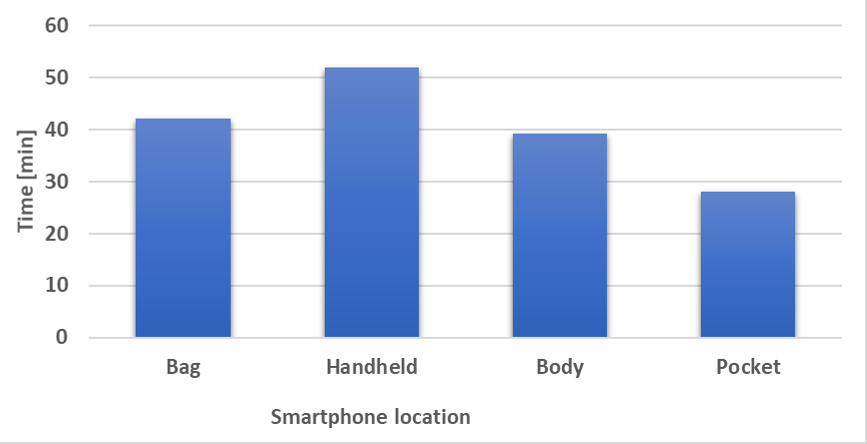}
    \caption{Amount of time for each of the four classes in the RIDI dataset.}
    \label{fig:ridi_class}
\end{figure}

\noindent The RIDI dataset is fully described in \cite{yan2018ridi} and is publicly accessible through Kaggle at: \url {https://www.kaggle.com/datasets/kmader/ridi-robust-imu-double-integration}. 
The dataset comprises IMU sensor measurements and motion trajectories from ten human subjects in four common smartphone positions: pocket,  bag,  handheld, and on the body. 
Over 160 minutes of data were collected at a sampling rate of 200 Hz encompassing various motion types such as walking forward/backwards, side motion, or acceleration/deceleration. The recordings were collected with a Google Tango phone, Lenovo Phab2 Pro.
In this work, we utilized the RIDI dataset to predict the smartphone location with four different classes as illustrated in Figure~\ref{fig:ridi_class}.
\subsubsection{MotionSense Dataset} \label{MS}
\begin{figure}[h]
    \centering
    \includegraphics[width=\columnwidth]{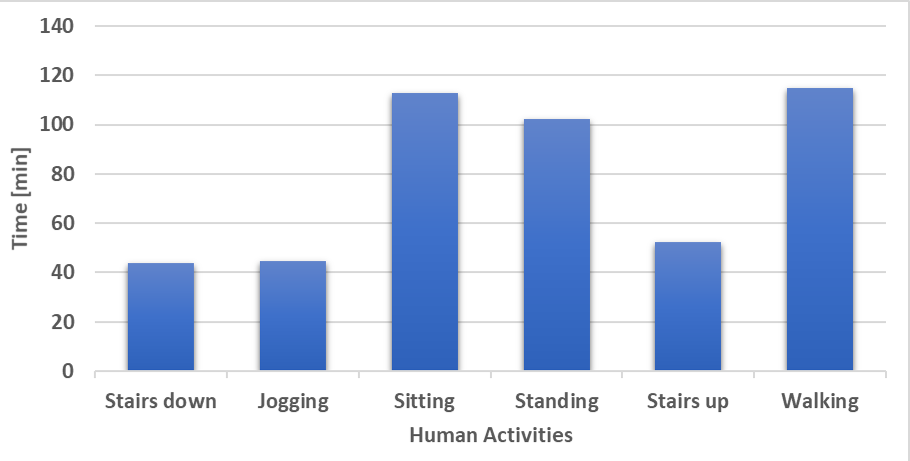}
    \caption{Amount of time for each of the six classes in the MotionSense dataset.}
    \label{fig:ms_class}
\end{figure}

The MotionSense dataset was collected with an iPhone 6s kept in the participant's front pocket with 24 participants (10 women and 14 men) with varying age, weight, and height. The  dataset is fully described in \cite{malekzadeh2019mobile} and is publicly accessible at: \url {https://github.com/mmalekzadeh/motion-sense}. 
The participants performed six activities:  Walking, Jogging, Sitting, Standing, Stairs down, and Stairs up as presented in Figure~\ref{fig:ms_class}. A total of 435 minutes of inertial recordings are provided at a sampling rate of 50Hz.
\subsubsection{HAR Dataset} \label{HAR1}
\begin{figure}[h]
    \centering
    \includegraphics[width=\columnwidth]{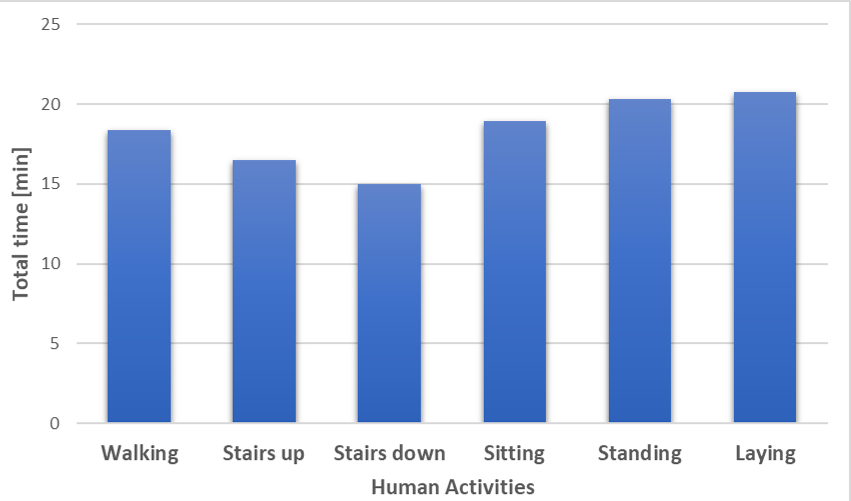}
    \caption{Amount of time for each of the six classes in the HAR dataset.}
    \label{fig:HAR_class}
\end{figure}

In this HAR dataset, the experiments have been carried out with a group of 30 participants within an age range of 19-48 years. Each person performed six activities:  Walking, Stairs down, Stairs up,  Sitting, Standing, and Laying as presented in Figure~\ref{fig:HAR_class}. In all of the activities the smartphone (Samsung Galaxy S II) was mounted on the waist. 
The  dataset is fully described in \cite{anguita2013public} and is publicly accessible at: \url{https://archive.ics.uci.edu/dataset/240/human+activity+recognition+using+smartphones}. A total of 110 minutes of inertial recordings are provided at a sampling rate of 50Hz.
\subsubsection{USC-SIPI HAR Dataset} \label{USCSIPI}
\begin{figure}[h]
    \centering
    \includegraphics[width=\columnwidth]{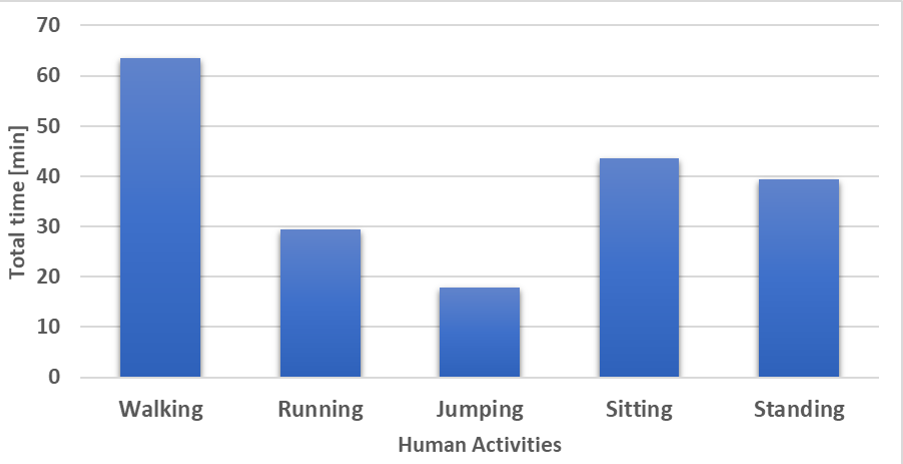}
    \caption{Amount of time for each of the six classes in the USC-SIPI HAR dataset.}
    \label{fig:UHAR_class}
\end{figure}

In this HAR dataset, the experiments have been carried out with a group of 14 participants (7 women and 7 men) within an age range of 21-49 years. They used an IMU firmly mounted into a mobile phone pouch and attached to the subject’s front right hip. 
The  dataset is fully described in \cite{zhang2012usc} and is publicly accessible at: \url{https://sipi.usc.edu/had/}. A total of 193 minutes of inertial recordings are provided at a sampling rate of 100Hz.  Each person performed five activities:  Walking, Running, Jumping,  Sitting, and Standing as presented in Figure~\ref{fig:UHAR_class}.
\section{Results} \label{sec:res}
\noindent
We utilized the accuracy measure as the evaluation metric to evaluate all datasets. The baseline accuracy classification results are given in Table~\ref{tbl:datasets_info}. In the next subsections, we present the results of the data-driven perspectives on each dataset and give a summary of the results.
\begin{table}[!thb]
    \centering
    \caption{Baseline accuracy results.}
    \label{tbl:datasets_info}
    %\begin{adjustbox}{width = 1.2\columnwidth}
    \begin{tabular}{cc}
    \hline
    \makecell[l]{\textbf{Dataset}} &  \makecell[l]{\textbf{Baseline Accuracy}}  \\ \hline
    \makecell[l]{RIDI } &  94.5  \\ \hline
    \makecell[l]{MotionSense } &  89.9  \\ \hline
    \makecell[l]{HAR } &  86.0  \\ \hline
    \makecell[l]{USC-SIPI HAR } &  95.4  \\ \hline
    \end{tabular}
   % \end{adjustbox}
\end{table}
\subsubsection{RIDI Dataset} 
\begin{figure}[h]
    \centering
    \includegraphics[width=\columnwidth]{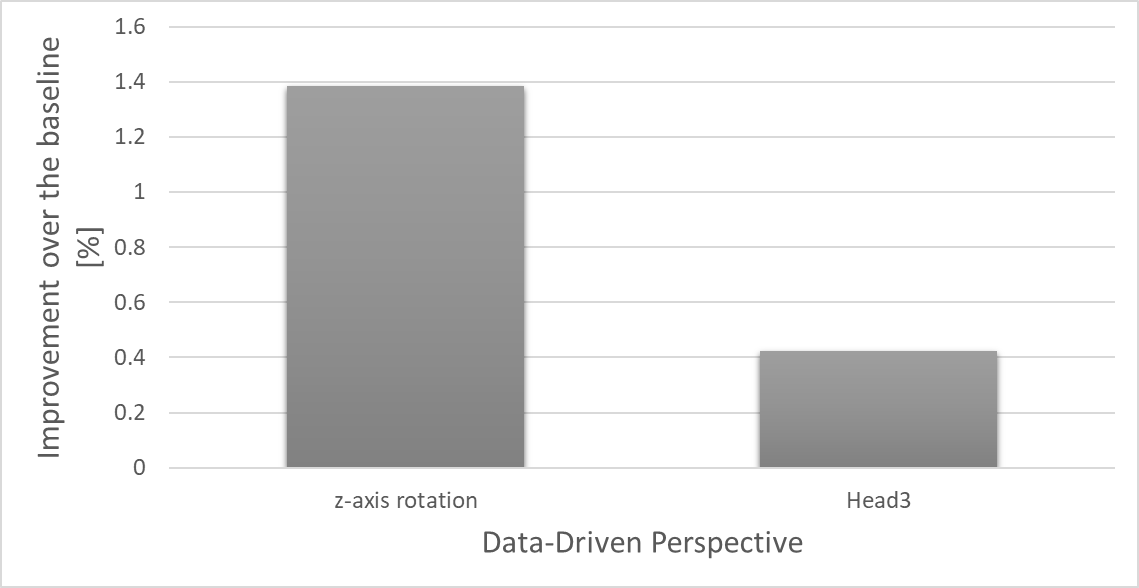}
    \caption{Improvements over the baseline accuracy on the RIDI dataset.}
    \label{fig:ridi_res}
\end{figure}

The baseline accuracy result for the RIDI dataset is $94.5\%$ posing a challenge to the possible improvement. Nevertheless, in two scenarios an improvement was achieved, where using the the z-axis rotation a maximum of $1.4\%$ was demonstrated as shown in Figure~\ref{fig:ridi_res}.
\subsubsection{MotionSense Dataset} 
\begin{figure}[h]
    \centering
    \includegraphics[width=\columnwidth]{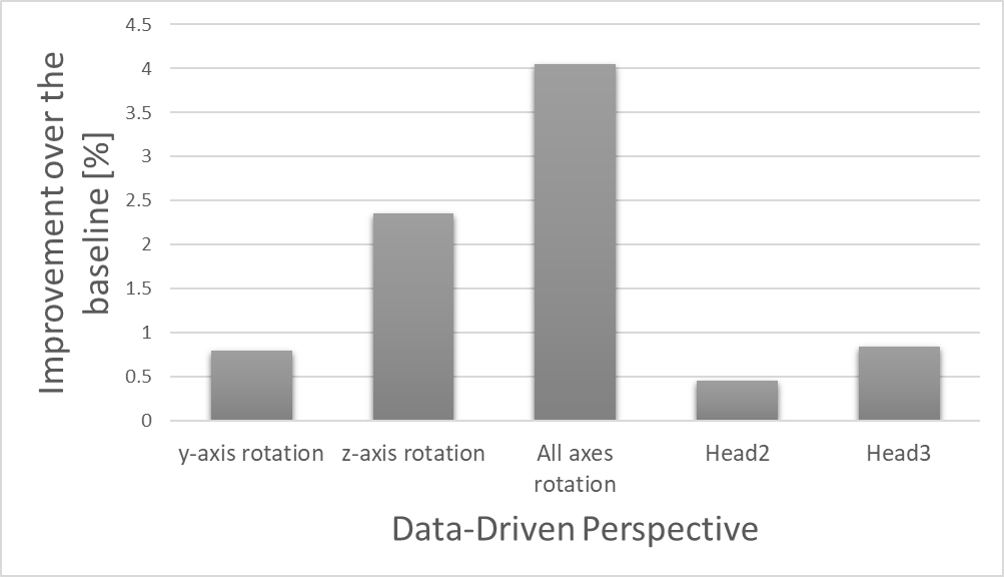}
    \caption{Improvements over the baseline accuracy on the MotionSense dataset.}
    \label{fig:ms_res}
\end{figure}

On this dataset 5/10 scenarios  improved the baseline results in the two types of multi-head architectures and 3/4 rotation options. A maximum improvement of  $4.0\%$ was achieved using all axes rotation, as presented in Figure~\ref{fig:ms_res}.
\subsubsection{HAR Dataset}
\begin{figure}[h]
    \centering
    \includegraphics[width=\columnwidth]{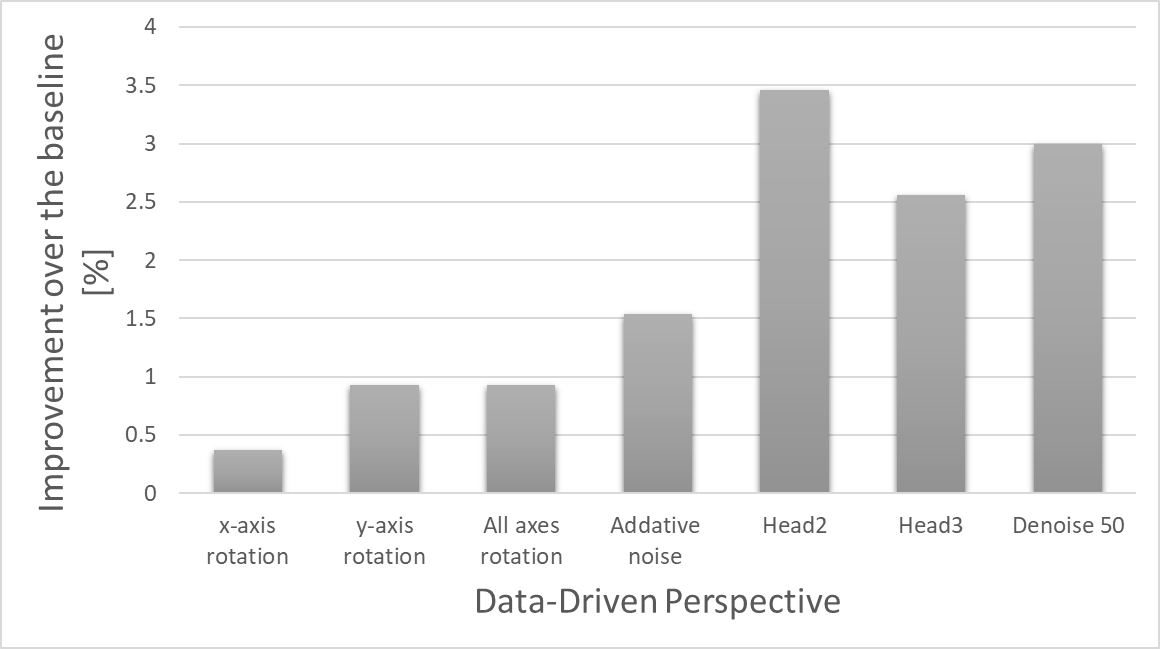}
    \caption{Improvements over the baseline accuracy on the HAR dataset.}
    \label{fig:HAR_res}
\end{figure}

In the HAR dataset seven out ten scenarios mange to improve the accuracy performance. It is notable that all types of data-driven perspectives improved performance including network architecture design, data augmentation, and data prepossessing techniques.  A maximum improvement of $3.5\%$ was obtained using all axes rotation approach, as presented in Figure~\ref{fig:HAR_res}.
\subsubsection{USC-SIPI HAR Dataset} 
\begin{figure}[h]
    \centering
    \includegraphics[width=\columnwidth]{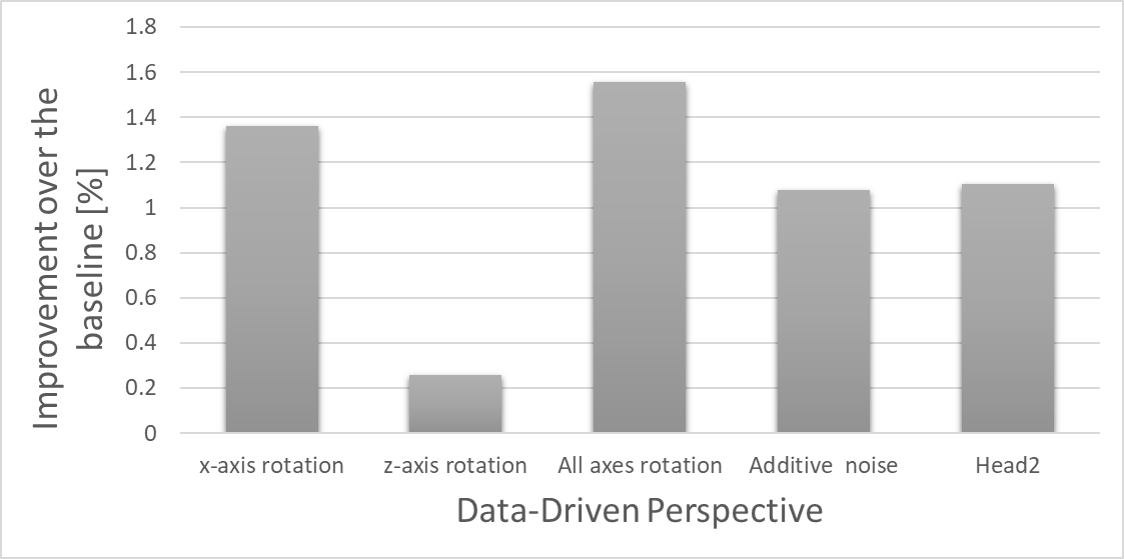}
    \caption{Improvements over the baseline accuracy on the USC-SIPI HAR dataset.}
    \label{fig:UHAR_res}
\end{figure}

The baseline accuracy result of the USC-SIPI HAR Dataset is $95.4\%$, which presents a challenge to further improvement. Yet, five of ten scenarios manged to improve the performance, as shown in Figure~\ref{fig:UHAR_res}.
\subsubsection{Summary} 
\begin{figure*}[h]
    \centering
    \includegraphics[width=\textwidth]{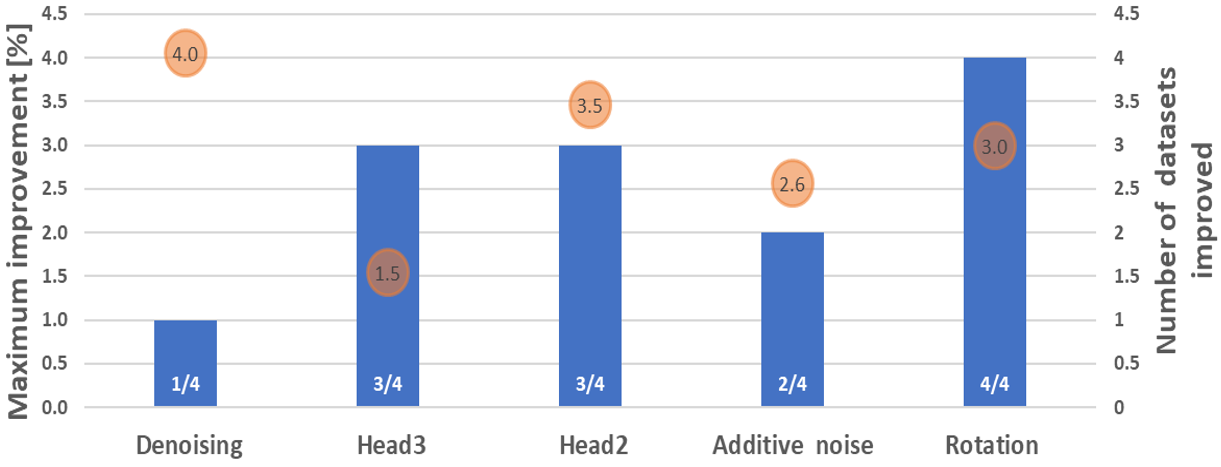}
    \caption{Summary of the results showing the number of datasets that were improved by applying data-driven techniques (blue bars) and the maximum improvement of each of the techniques (orange circles).}
    \label{fig:Res_ALL}
\end{figure*}

A summary of the main results on all four datasets and ten experiments is presented in Figure~\ref{fig:Res_ALL}.  The figure presents number of datasets that were improved by applying data-driven techniques and the maximum improvement of each of the techniques. The most effective techniques for improving accuracy were data augmentation through rotation (4/4 datasets) and noise multi-head architectures (3/4 datasets). As the baseline accuracy results are very high and in the range of $86.0-95.4\%$, improving the performance was challenging. Nevertheless, the proposed approaches manged to improve all datasets with a number of successful techniques ranging from two to seven (out of ten). The maximum improvement for each technique varied from $1.5\%$ to $4.0\%$.
\section{Conclusions} \label{sec:conclusions}
\noindent 
This research presented a comprehensive evaluation of ten techniques for improving deep learning models' accuracy applied to inertial classification networks focusing on pedestrian activity recognition problems. Three different aspects were explored: network architectural design, data augmentation, and data preprocessing across four datasets. Those datasets include 936 minutes of inertial data sampled between 50-200Hz and recorded by 78 participants. Our findings demonstrate that data augmentation methods through rotation and multi-head architectures emerged as the most effective in improving performance across datasets. 
\\
In conclusion, this study presents effective strategies to improve inertial classification networks' accuracy for pedestrian activity recognition. It underscores the importance of dataset-specific tuning. Additionally, this study outlines benchmarking strategies for enhancing neural inertial regression networks.
\section*{Acknowledgment}
Z. Y. is supported by the Maurice Hatter Foundation.

\bibliographystyle{IEEEtran}
\bibliography{./main.bib}

\end{document}